\documentclass[sigconf]{acmart}
\AtBeginDocument{%
  }

\setcopyright{acmlicensed}
\copyrightyear{2024}
\acmYear{2024}
\acmDOI{XXXXXXX.XXXXXXX}
\acmConference[Conference acronym 'XX]{Make sure to enter the correct
  conference title from your rights confirmation email}{June 03--05,
  2018}{Woodstock, NY}

\renewcommand\footnotetextcopyrightpermission[1]{} 

\acmISBN{978-1-4503-XXXX-X/18/06}

\usepackage{multirow}
\newcommand{\eg}[1]{\textit{e.g.,}}
\newcommand{\ie}[1]{\textit{i.e.,}}
\begin{document}
\fancyhf{} %
\pagestyle{empty} %
\title{TMCIR: Token Merge Benefits Composed Image Retrieval}

\author{Chaoyang Wang$^1$\quad Zeyu Zhang$^{2*}$\quad Long Teng$^3$\quad Zijun Li$^4$\quad Shichao Kan$^{1\dag}$\\
\vspace{0.2cm}
$^1$Central South University\quad
$^2$The Australian National University\\
$^3$Wuhan University\quad
$^4$Tsinghua University\\
\small $^*$Project lead. $^\dag$Corresponding author: kanshichao@csu.edu.cn.}


\begin{abstract}
Composed Image Retrieval (CIR) retrieves target images using a multi-modal query that combines a reference image with text describing desired modifications. The primary challenge is effectively fusing this visual and textual information. Current cross-modal feature fusion approaches for CIR exhibit an inherent bias in intention interpretation. These methods tend to disproportionately emphasize either the reference image features (visual-dominant fusion) or the textual modification intent (text-dominant fusion through image-to-text conversion). Such an imbalanced representation often fails to accurately capture and reflect the actual search intent of the user in the retrieval results.
To address this challenge, we propose TMCIR, a novel framework that advances composed image retrieval through two key innovations: 1) Intent-Aware Cross-Modal Alignment. We first fine-tune CLIP encoders contrastively using intent-reflecting pseudo-target images, synthesized from reference images and textual descriptions via a diffusion model. This step enhances the encoder ability of text to capture nuanced intents in textual descriptions. 2) Adaptive Token Fusion. We further fine-tune all encoders contrastively by comparing adaptive token-fusion features with the target image. This mechanism dynamically balances visual and textual representations within the contrastive learning pipeline, optimizing the composed feature for retrieval. Extensive experiments on Fashion-IQ and CIRR datasets demonstrate that TMCIR significantly outperforms state-of-the-art methods, particularly in capturing nuanced user intent.
\end{abstract}




\maketitle

\section{Introduction}
\begin{figure}[t]
\centering
\includegraphics[width=\linewidth]{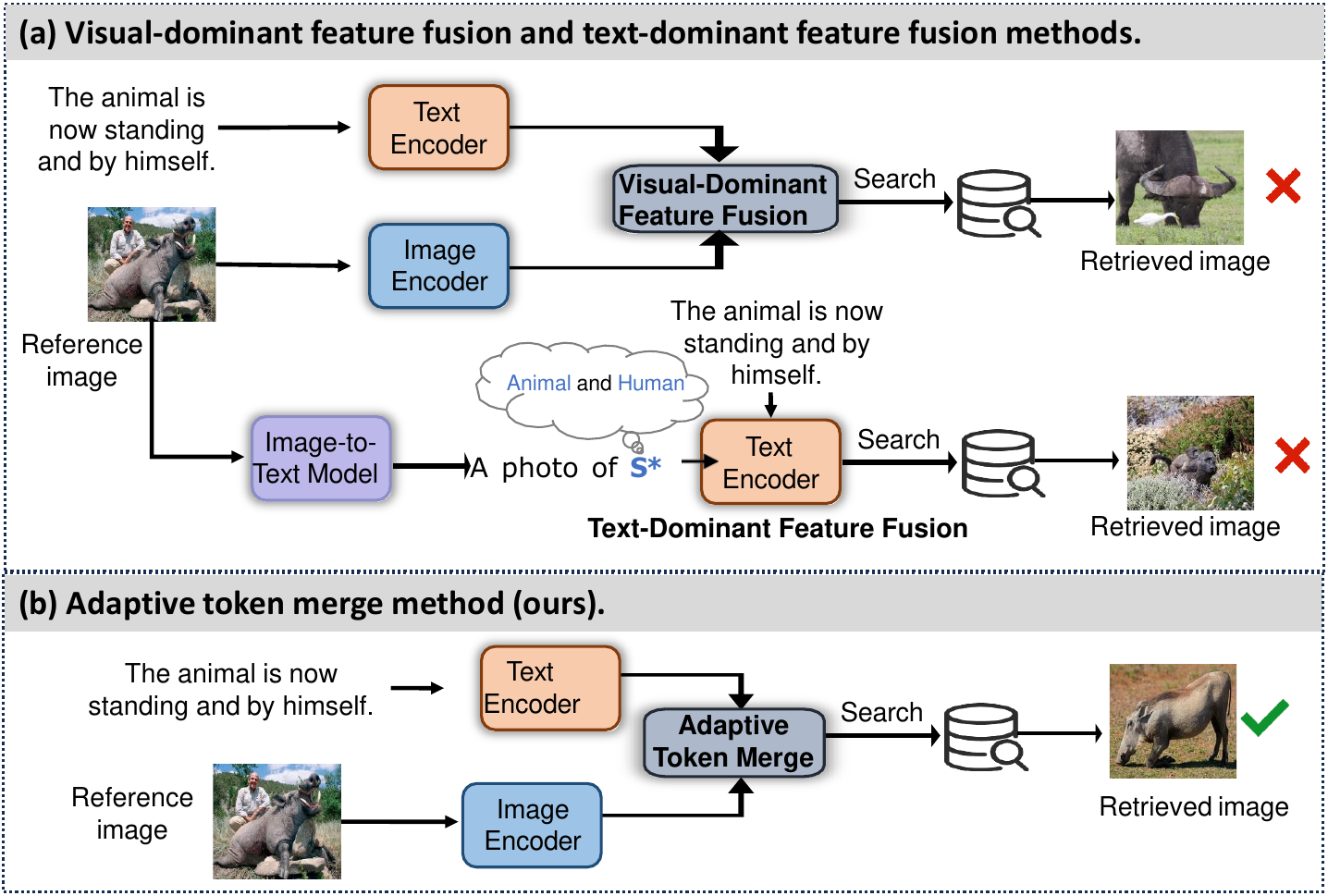}
\caption{Workflows of existing CIR methods and our proposed TMCIR}
\label{fig:intro}
\vspace{-0.5cm}
\end{figure}
Retrieving images based on a combination of a reference image and textual modification instructions defines the task of Composed $\sim$Image Retrieval (CIR) ~\citep{Lee05,vo2019composing,baldrati2022effective}. Specifically, the goal of CIR is to retrieve a target image from a candidate set that maintains overall visual similarity to the reference image while fulfilling the localized modification requirements specified in the textual description. CIR enables precise, interactive retrieval, making it valuable for applications like e-commerce and personalized web search.

However, composed queries from two distinct modalities introduces unique challenges. Unlike text-to-image or image-to-image retrieval which rely on a single query type, CIR must interpret \textit{relative changes} described textually and apply them to the \textit{specific visual content} of the reference image. The core difficulty lies in effectively integrating these cross-modal signals into a unified representation for similarity comparison with candidate images. To achieve this integration, most current approaches predominantly employ visual-dominant feature fusion mechanisms ~\citep{anwaar2021compositional,chen2020learning,liu2021image,dodds2020modality}.
which extract image and text features separately and then combining them. However, these methods exhibit two critical limitations: 1) They often fail to preserve essential visual details from the reference image; 2) They tend to inadvertently incorporate irrelevant background information (\ie, regions unrelated to the textual modifications) into the final query representation. These shortcomings become particularly pronounced in scenarios requiring fine-grained image modifications, such as precise color variations or localized texture alterations, where maintaining both visual fidelity and modification accuracy is paramount.

Other recent approaches have adopted text-dominant fusion mechanisms that leverage CLIP-based image-to-text conversion \cite{gal2022image,baldrati2022effective}, where reference images are mapped to pseudo-word embeddings for integration with textual descriptions. While this paradigm benefits from established cross-modal alignment, it faces fundamental limitations: 1) The generated pseudo-word tokens primarily capture global image semantics while losing fine-grained visual details; 2) The constrained length of the word tokens restricts comprehensive visual representation. These factors lead to granular-level discrepancies in the cross-modal representations, ultimately compromising retrieval accuracy.

As shown in Fig.~\ref{fig:intro}(a), both the visual-dominant fusion and the text-dominant fusion methods are fail to accurately capture and reflect the actual search intent of the user in the retrieval results. To address these challenges, we propose a TMCIR framework. Our framework is carefully designed to preserve the critical visual information present in the reference image while accurately conveying the modification intent of the user as specified by textual description. The TMCIR comprises two key steps:
1) \textbf{Intent-Aware Cross-Modal Alignment:} This step is designed to precisely capture textual intent from descriptions, addressing a critical limitation in CIR. Conventional target images often contain extraneous variations (\eg, lighting conditions, irrelevant background objects, or stylistic inconsistencies) that deviate from the specified intent of text, making them suboptimal for fine-tuning. To overcome this, we introduce a pseudo-target generation module that leverages a diffusion model conditioned on both the reference image and the relative textual description. The generated pseudo-target image eliminates noise, serving as a cleaner supervisory signal that faithfully reflects the intended modifications. Using an image-text paired dataset constructed from these pseudo-targets and their corresponding descriptions, we perform contrastive fine-tuning of pre-trained visual and textual encoders. This approach ensures precise cross-modal token alignment in a shared embedding space, with a focused emphasis on text-intent preservation from the description.
2) \textbf{Adaptive Token Fusion}: Following alignment fine-tuning, we introduce an adaptive fusion strategy that computes token-wise cosine similarity between visual and textual encoder outputs, enhanced with positional encoding for weighted feature fusion (Fig.~\ref{fig:intro}(b). This design serves two key purposes: 1) The positional cues establish explicit correspondences between textual concepts and their spatial counterparts in the image. 2) The similarity-weighted fusion preserves critical visual details while precisely encoding the nuanced modification intents specified in relative descriptions. The fused representations then drive a final contrastive fine-tuning stage, where we optimize all encoders by comparing the adaptive token-fusion features against target images. This dynamic balancing mechanism simultaneously refines both modalities within a unified contrastive framework, ultimately producing composite features that are optimally discriminative for retrieval tasks.

\begin{figure}[t]
\centering
\includegraphics[width=\linewidth]{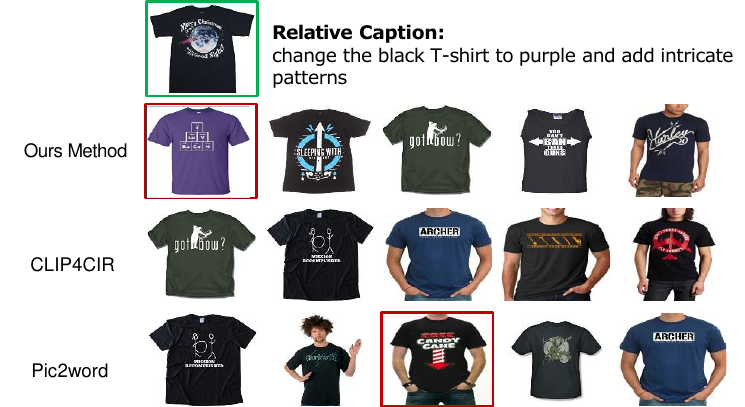}
\caption{Retrieval examples using the proposed TMCIR, CLIP4CIR \cite{baldrati2022conditioned} (visual-dominant feature fusion), and Pic2word \cite{saito2023pic2word} (text-dominant fusion) methods, respectively.}
\label{fig2}
\vspace{-0.5cm}
\end{figure}

Experiments conducted on the Fashion-IQ and CIRR datasets demonstrate that our method achieves state-of-the-art performance on both benchmarks. As shown in Figure \ref{fig2}, a user provides a reference image of a black T-shirt accompanied by the relative description ``change the black T-shirt to purple and add intricate patterns''. The method of CLIP4CIR\cite{baldrati2022conditioned} employs visual-dominant feature fusion, which risks incorporating irrelevant visual elements (e.g., background clutter) that dilute the precise modification intent, leading to erroneous results. Conversely, the method of Pic2word\cite{saito2023pic2word} uses text-dominant fusion that may suppress relevant visual details while similarly weakening the intended modifications, resulting in comparable failures. Both approaches highlight the need for a balanced fusion strategy that preserves critical visual features while faithfully capturing textual intentions. Our adaptive token fusion approach analyzes the image in finer detail. It partitions the reference image, allowing visual tokens from different regions to be compared with keywords in the description. The mechanism can distinguish the T-shirt region from the background and, via cosine similarity, identify tokens highly correlated with keywords like ``purple'', ``T-shirt'', and ``patterns''. Tokens from the T-shirt region receive higher weights, while background tokens are attenuated. Simultaneously, the textual information for ``purple'' guides the adjustment of the token representations associated with the T-shirt, generating a joint representation that accurately fulfills the modification requirements.

In summary, our contributions are as follows:
\begin{itemize}
\item We propose a novel CIR approach that integrates intent-aware cross-modal alignment (IACMA) and adaptive token fusion (ATF) to better capture user intent. The IACMA leverages a diffusion model to generate pseudo-target images that more accurately reflect user modification intent compared to potentially noisy real target images, providing a purer supervisory signal for encoder fine-tuning.
\item The ATF adaptively fuse visual and textual tokens through weighted integration and positional encoding. This strategy ensures comprehensive preservation of key visual details while accurately capturing subtle user modification intent.
\item Experimental results on the Fashion-IQ and CIRR datasets indicate that our proposed method outperforms current state-of-the-art CIR approaches in both retrieval accuracy and robustness.
\end{itemize}

\section{Related work}
\textbf{Composed Image Retrieval.} 
%
In current mainstream CIR methods, one category focuses on the reference image features, where the features of the reference image are fused with those of the relative caption. 
Then the fused feature is compared with those of all candidate images to determine the closest match~\citep{anwaar2021compositional,chen2020image,dodds2020modality,liu2021image,vo2019composing}. 
Various feature fusion mechanisms~\citep{vo2019composing} and attention mechanisms~\citep{chen2020image,dodds2020modality} have exhibited prominent performance in CIR. 
Subsequently, capitalizing on the robust capabilities of pre-trained models, a number of CIR methods that adeptly amalgamate image and text features extracted from autonomously trained visual and text encoders~\citep{baldrati2022conditioned,goenka2022fashionvlp,ray2023cola}. 
%
%
%
%
Another category of approaches suggest to transform reference image into its pseudo-word embedding~\citep{saito2023pic2word,baldrati2023zero}, which is then combined with relative caption for text-to-image retrieval.
However, the pseudo-word embeddings learned in this manner often capture only global semantic information while lacking fine-grained visual details, and their limited length makes it difficult to comprehensively represent the visual content of the reference image.
%
Liu \textit{et al.} created text with semantics opposite to that of the original text and add learnable tokens to the text to retrieve images in two distinct queries~\citep{liu2023candidate}.
Nevertheless, this approach to fixed text prompt learning fails to modify the intrinsic information within the relative caption itself, thereby constraining retrieval performance. 
Although existing composed image retrieval methods have achieved varying degrees of visual and textual information fusion, they still fall short in fully preserving the detailed information in the reference image and accurately aligning with the user's modification intent. Motivated by this, we propose the TMCIR framework, which effectively addresses the issues of information loss and insufficient cross-modal alignment through the generation of pseudo-target images, task-specific fine-tuning of encoders, and a similarity-based token fusion module. As a result, our approach achieves significant improvements in both retrieval accuracy and robustness.

\textbf{Token Merge for Modal Fusion.}
Multi-modal fusion requires the compression of redundant features and efficient representation learning, both of which are critical for enhancing a model’s generalization and inference efficiency. Traditional multi-modal fusion approaches often suffer from redundant features and escalated computational costs due to overlapping information across modalities, particularly in resource-constrained settings. The Token Merge method was originally proposed for vision Transformers to reduce unnecessary computation by merging redundant or semantically similar tokens \cite{bolya2022token}. Subsequently, this method was adapted to multi-modal tasks where researchers attempted to dynamically fuse tokens from different modalities to construct a more compact and expressive joint representation. In tasks such as image–text retrieval, image–text generation, and video semantic understanding, Token Merge not only improves computational efficiency but also significantly alleviates the issues of information conflicts and semantic inconsistencies during cross-modal alignment \cite{chen2024evlm,luo2023cheap,liu2023token,shen2023scaling}.

In the context of composed image retrieval, the Token Merge method is used to address the inconsistent granularity between cross-modal feature alignment and fusion. For example, Wang et al. \cite{wang2023agree} emphasized cross-modal entity alignment during fine-tuning by employing contrastive learning and entity-level masked modeling to promote effective information interaction and structural unification between image and text. To further exploit the potential of localized textual representations, NSFSE \cite{wang2024negative} adaptively delineated the boundaries between matching and non-matching samples by comprehensively considering the correspondence of positive and negative pairs.

Furthermore, efficient inference and generation in multi-modal models can also be achieved by directly pruning labels in the language model. Huang \cite{huang2024prunevid} introduced a training-free method that minimizes video redundancy by merging spatio-temporal tokens, thereby enhancing model efficiency. Compared with traditional attention-intensive interaction mechanisms, Token Merge offers a more flexible and controllable computational pathway for multimodal fusion, with promising scalability and practical engineering prospects.

In our work, we apply the Token Merge technique to the fusion of visual and textual information in composed image retrieval. By adaptively merging visual tokens and textual tokens, our approach better preserves essential visual details from the reference image and fine-grained semantic information from the text. This results in more precise cross-modal token alignment and fusion, ultimately enhancing the retrieval performance.

\begin{figure*}[ht!]
\centering
\includegraphics[width=1\linewidth]{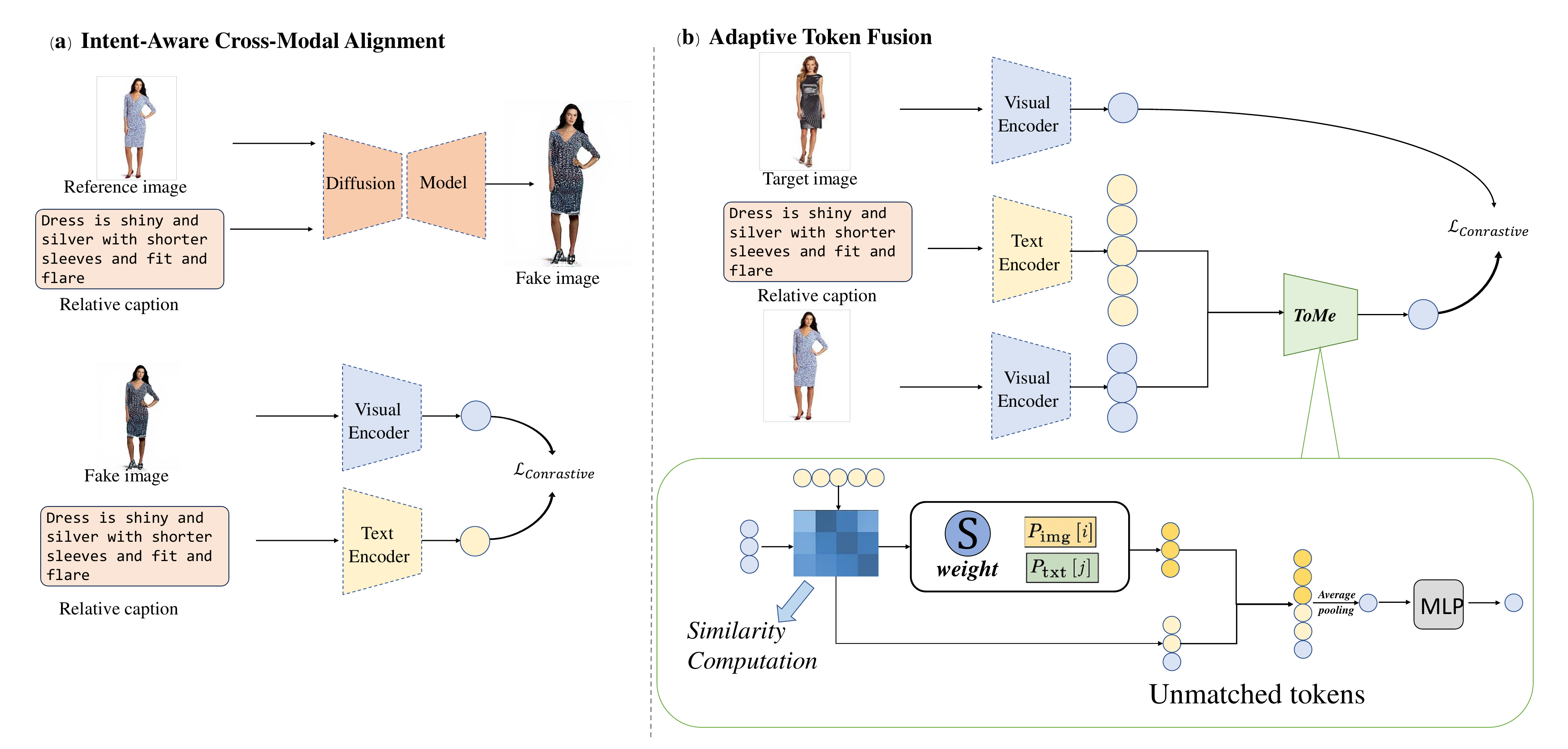}
\caption{An Overview of the TMCIR Framework.It consists of two modules: the "Intent-Aware Cross-Modal Alignment" module and the "Adaptive Token Fusion" module. First, we input the reference image and the relative description into a diffusion model to generate a pseudo-target image. Through contrastive learning, we guide the visual and textual encoders to achieve cross-modal token distribution alignment. Then, the reference image and the relative description are fused using an adaptive token fusion strategy based on positional encoding and semantic similarity, generating a joint representation that captures both the user intent and the key visual information from the reference image.}
\Description[Overview]{Overview of Our Framework of Scaling Positive and Negative Examples.}
\label{fig:overview}
\end{figure*}

\section{Method}
\subsection{Preliminary}
Assume that a composed image retrieval (CIR) dataset contains $N$ annotated triplet samples, where the $i$th triplet sample $x_i$ is represented as:
\begin{equation}
x_i = (r_i, m_i, t_i), \quad r_i, t_i \in \Omega, \quad m_i \in \mathcal{T}.
\end{equation}
Here, $r_i$, $m_i$, and $t_i$ denote the reference image, the \emph{relative} description, and the target image of the $i$th triplet sample, respectively. The term \emph{relative} emphasizes that $m_i$ specifies the modifications to be applied to $r_i$ to obtain the target image, capturing how the target image differs from the reference image. $\Omega$ represents the candidate image set that contains all reference and target images, and $\mathcal{T}$ denotes the text set containing all relative descriptions.

In the CIR task, the query $q_i$, which is composed of the reference image $r_i$ and the relative description $m_i$, is used to retrieve the target image $t_i$ from the candidate set $\Omega$. In the classical CIR training paradigm, multiple annotated triplet samples are first grouped into a mini-batch. Within the same batch, the reference images and relative descriptions are encoded into query representations by a query encoder $F(\cdot)$, while the target images are encoded by an image encoder $G(\cdot)$ to obtain their embeddings. For brevity, we denote the representations of the triplet $(r_i, m_i, t_i)$ as
\[
q_i = F(r_i, m_i) \quad \text{and} \quad v_i = G(t_i),
\]
where $v_i$ denotes the embedding of the target image. The cosine similarity function $f(\cdot, \cdot)$ computes the similarity between the query representation and the target image embedding. Most current methods adopt a contrastive learning paradigm, which pulls together the query and target image representations of positive pairs (i.e., the query paired with its matching target image) while pushing apart those of negative pairs (i.e., a query paired with a target image from a different triplet). The corresponding loss function is formulated as:
\begin{equation}
L_{cl} = \frac{1}{B} \sum_{i=1}^{B} - \log \left( \frac{\exp(f(q_i, v_i) / \tau)}{\sum_{j=1}^{B} \exp(f(q_i, v_j) / \tau)} \right),
\end{equation}
where $B$ is the batch size and $\tau$ is the temperature hyperparameter, which controls the sharpness of the similarity distribution and thus regulates the strength of the contrastive signal.

Our method follows the same contrastive learning paradigm.

\subsection{Overview}
\label{bbb}
As depicted in Figure~1, our proposed TMCIR pipeline comprises two steps:

(1) \textbf{Intent-Aware Cross-Modal Alignment.} This step contains \textbf{Pseudo Target Image Generation} (PTIG) and \textbf{Encoder Fine-Tuning for Token Alignment} (EFTTA) modules. In the PTIG module, a diffusion model, specifically Stable Diffusion~3, is utilized to generate a pseudo target image $p_i$ by conditioning on the reference image $r_i$ and the relative description $m_i$. This pseudo target image accurately reflects the modification requirements specified in $m_i$, ensuring a high level of controllability and reproducibility.
In the EFTTA module, we construct an image-text pair dataset from the relative description and the pseudo target image, then fine-tune the visual and text encoders of the CLIP model. This process promotes more consistent cross-modal token distributions in the shared embedding space. Here, \emph{token alignment} refers to the process of harmonizing tokens from different modalities in the embedding space so that their semantic representations and attention patterns become more correlated and comparable.

(2) \textbf{Adaptive Token Fusion.} After obtaining visual and text tokens from the fine-tuned bimodal encoder, we design an adaptive token fusion strategy. In our approach, token merging is performed on a token-by-token basis by computing the cosine similarity between individual tokens and incorporating positional encoding via weighted averaging. This fusion strategy constructs a unified and semantically rich cross-modal representation.

In the following subsections, we provide details for these two steps.

\subsection{Intent-Aware Cross-Modal Alignment}
The Intent-Aware Cross-Modal Alignment step aims to enhance the encoder ability of text to capture nuanced intents in textual descriptions, which includes pseudo target image generation and encoder fine-tuning for token alignment modules.

\paragraph{\textbf{Pseudo Target Image Generation}}
\label{ccc}
Benefiting from large-scale training data comprising billions of image-text pairs, large-scale visual-language models (\eg, CLIP\citep{radford2021learning}) have demonstrated excellent generalization capabilities across numerous downstream tasks. This has inspired the application of such foundational models to the composite image retrieval (CIR) task. Existing methods typically rely on various modality fusion strategies to integrate the bimodal features extracted by the visual and text encoders. However, the visual encoder in pre-trained models primarily focuses on overall visual information, while the text encoder captures generic language features. Directly employing the original encoders often leads to inconsistent token distributions between the visual and text modalities in specific CIR datasets. This inconsistency results in suboptimal similarity computations during the token fusion phase, and consequently, inferior retrieval performance.

To address this issue, we first select image-text pairs from existing CIR datasets---typically sampling all available pairs or a fixed number per batch in our experiments---and perform task-specific fine-tuning of the visual and text encoders to achieve more consistent token representations from the reference image and the relative description. Considering that the manually collected triplet samples in current CIR datasets contain target images sourced from diverse origins (which may include background interference or noise not aligned with the modification description), we generate a pseudo target image $p_i$ by conditioning a diffusion model $D$ on the reference image $r_i$ and the relative description $m_i$: 
\begin{equation}
p_i = D(r_i, m_i).
\end{equation}
The pseudo target image $p_i$ accurately embodies the modification requirements stipulated in $m_i$, while excluding irrelevant background noise. This provides a purer and more precise supervisory signal for subsequent encoder fine-tuning. The pseudo target image $p_i$ and the relative description $m_i$ are then combined to form an image-text pair dataset $\mathcal{D}$:
\begin{equation}
\mathcal{D} = \{ (m_i, p_i) \mid i = 1, 2, \dots, N \}.
\end{equation}
We utilize $\mathcal{D}$ as a dedicated dataset in a distinct training stage for fine-tuning the encoders, separate from the original CIR training set.

\paragraph{\textbf{Encoder Fine-Tuning for Token Alignment}}
After constructing the image-text pair dataset $\mathcal{D} = \{(m_i, p_i)\}_{i=1}^N$ as described in Section~\ref{ccc}, we adopt a contrastive learning strategy to fine-tune the CLIP-pretrained visual encoder and text encoder, thereby further enhancing their cross-modal representation and alignment abilities for the CIR task.

Specifically, for each image-text pair $(m_i, p_i)$, the relative description $m_i$ is input into the text encoder $E_T$ to obtain the text feature vector:
\begin{equation}
t_i = E_T(m_i)    
\end{equation}

while the pseudo target image $p_i$ is input into the visual encoder $E_V$ to obtain the image feature vector:
\begin{equation}
v_i = E_V(p_i)
\end{equation}

We then normalize these feature representations:
\begin{equation}
\hat{t}_i = \frac{t_i}{\|t_i\|_2}, \quad \hat{v}_i = \frac{v_i}{\|v_i\|_2}    
\end{equation}

to facilitate subsequent cosine similarity calculations.

Next, we optimize the model parameters using the InfoNCE loss function:
\begin{equation}
\mathcal{L}_{\text{}} = -\frac{1}{N} \sum_{i=1}^{N} \log \frac{\exp(\text{sim}(\hat{v}_i, \hat{t}_i) / \tau)}{\sum_{j=1}^{N} \exp(\text{sim}(\hat{v}_i, \hat{t}_j) / \tau)}    
\end{equation}

where $\text{sim}(\cdot, \cdot)$ denotes the cosine similarity function and $\tau$ is a learnable temperature parameter that controls the smoothness of the similarity distribution.

Following fine-tuning, the token representations produced by the visual and text encoders in the shared embedding space exhibit improved distribution consistency and cross-modal alignment, thereby providing a robust foundation for the subsequent token merging module.

\subsection{Adaptive Token Fusion}
Following the fine-tuning of the visual and text encoders, the resulting visual tokens and text tokens are well-aligned in the shared embedding space. To effectively integrate the dual-modal information, we design a token merging module based on similarity computation.

Specifically, the reference image $r_i$ and the relative description $m_i$ are input into the fine-tuned visual encoder $E_V$ and text encoder $E_T$, respectively, to obtain the corresponding sets of tokens:
\begin{equation}
V = \{v_1, v_2, \dots, v_L\} = E_V(r_i),
\end{equation}
\begin{equation}
T = \{t_1, t_2, \dots, t_M\} = E_T(m_i).
\end{equation}
We then compute the similarity matrix $\mathbf{S} \in \mathbb{R}^{L \times M}$ between the image token set $V$ and the text token set $T$:
\begin{equation}
\mathbf{S}_{ij} = \frac{\mathbf{v}_i \cdot \mathbf{t}_j}{\|\mathbf{v}_i\| \cdot \|\mathbf{t}_j\|}.
\end{equation}
For each text token, we iterate over each visual token, considering a pair $(\mathbf{v}_i, \mathbf{t}_j)$ as a valid matching token pair if their similarity exceeds a preset threshold $\tau$.

\textbf{Fusion Strategy:} For each matching token pair $(\mathbf{v}_i, \mathbf{t}_j)$, we compute a weighted average using their similarity coefficient $\lambda$ as the weight and integrate their positional encodings ($\mathbf{P}_\text{img}[i]$ and $\mathbf{P}_\text{txt}[j]$) to preserve spatial information. The resulting fused representation $\mathbf{f}_{i,j}$ is calculated as:
\begin{equation}
\mathbf{f}_{i,j} = \frac{\mathbf{S}_{ij} \cdot \mathbf{v}_i + \mathbf{S}_{ij} \cdot \mathbf{t}_j}{2\mathbf{S}_{ij} + \epsilon} + 0.5 \cdot \left(\mathbf{P}_\text{img}[i] + \mathbf{P}_\text{txt}[j]\right),
\end{equation}
where $\epsilon$ is a small constant to prevent division by zero.

For visual tokens and text tokens that do not find a matching counterpart, we retain them by directly adding their positional residuals:
\begin{equation}
\tilde{\mathbf{v}}_i = \mathbf{v}_i + 0.5 \cdot \mathbf{P}_\text{img}[i], \quad \text{if } \mathbf{v}_i \notin \text{Matched},
\end{equation}
\begin{equation}
\tilde{\mathbf{t}}_j = \mathbf{t}_j + 0.5 \cdot \mathbf{P}_\text{txt}[j], \quad \text{if } \mathbf{t}_j \notin \text{Matched}.
\end{equation}
Finally, we concatenate all the fused tokens and the remaining unmatched tokens to form the cross-modal token sequence $\mathbf{Z}$:
\begin{equation}
    \mathbf{Z} = \left\{ \mathbf{f}_{ij} \mid (i, j) \in \mathcal{M} \right\} \cup \left\{ \tilde{\mathbf{v}}_i \mid i \notin \mathcal{M}_I \right\} \cup \left\{ \tilde{\mathbf{t}}_j \mid j \notin \mathcal{M}_T \right\}
\end{equation}

where $\mathcal{M}$ denotes the set of matching pairs, while $\mathcal{M}_I$ and $\mathcal{M}_T$ represent the matching indices for visual and text tokens, respectively. We then apply average pooling to $\mathbf{Z}$ to obtain a single token representation $\mathbf{z}$:
\begin{equation}
\mathbf{z} = \frac{1}{N_Z} \sum_{n=1}^{N_Z} \mathbf{Z}_n,
\end{equation}
and pass it through a fully connected layer $\mathrm{F}$ to obtain the final cross-modal embedding vector $V_Q$:
\begin{equation}
V_Q = \mathrm{F}(\mathbf{z}).
\end{equation}

\textbf{Learning Objective:} Our training objective for the composed image retrieval (CIR) task is to align the joint feature representation $V_Q$ of the mixed-modal query $(r, m)$ with the feature representation $V_T$ of the target image $t$. In each training iteration, we process a mini-batch of samples:
\begin{equation}
\{(V_Q^{(i)}, V_{T}^{(i)})\}_{i=1}^{N_B},
\end{equation}
where $(V_Q^{(i)}, V_{T}^{(i)})$ denotes the feature representations of the $i$th (mixed-modal query, target image) pair, and $N_B$ is the mini-batch size. The batch-based classification loss function is defined as:
\begin{equation}
L = \frac{1}{N_B} \sum_{i=1}^{N_B} -\log \frac{\exp(\lambda \cdot \text{Sim}(V_Q^{(i)}, V_{T}^{(i)}))}{\sum_{j=1}^{NB} \exp(\lambda \cdot \text{Sim}(V_Q^{(j)}, V_{T}^{(j)}))},
\end{equation}
where $\text{Sim}(\cdot)$ denotes the cosine similarity function and $\lambda$ is a temperature parameter.
\section{Experiment}
\subsection{Experimental Setup}
\textbf{Implementation Details.}
Our method is implemented in PyTorch and runs on an NVIDIA RTX A100 GPU with 80GB of memory. We adhere to the design principles of CLIP, initializing both the visual and text encoders from a CLIP pre-trained model based on the ViT-L architecture. The AdamW optimizer \cite{loshchilov2017decoupled} is employed with a weight decay coefficient set to 0.05. Input images are resized to 224×224 pixels, and a padding ratio of 1.25 is applied for uniform processing \cite{baldrati2022effective}. The initial learning rates are set to 1e-5 and 2e-5 for the CIRR and Fashion-IQ datasets, respectively, and a cosine learning rate scheduling strategy is adopted. The similarity threshold $\tau$ is set to 0.7.In the Pseudo-Target Generation stage, the diffusion model FLUX.1-dev-edit-v0 is utilized.

\textbf{Datasets and Metrics.} We evaluate our method on two CIR benchmarks: \texttt{(1)} \textbf{Fashion-IQ} a fashion dataset with $77,684$ images forming $30,134$ triplets~\cite{wu2021fashion}.We utilize Recall@K as the evaluation metric, which reflect the percentage of queries whose true target ranked within the top $K$ candidates.Since the ground-truth labels for the test set of this dataset have not been publicly disclosed, we adopt the results on the validation set for performance evaluation.
\texttt{(2)} \textbf{CIRR} is a general image dataset that comprises $36,554$ triplets derived from $21,552$ images from the popular natural language inference dataset NLVR2~\cite{suhr2018corpus}.
We randomly split this dataset into \texttt{training}, \texttt{validation}, and \texttt{test} sets in an $8:1:1$ ratio. 
This dataset encompasses rich object interactions, addressing the issues of overly narrow domains and high number of false-negatives in the \textbf{Fashion-IQ} dataset, thereby allowing for a comprehensive evaluation of the effectiveness of our proposed method.
We report the results of the competing methods on this dataset at different levels, \ie, Recall@1, 5, 10, 50, and Recallsubset@K~\citep{liu2021image}.

\begin{table}[h]
\centering
\caption{Ablation studies with regard to the impact of using pseudo versus real target images on retrieval performance.}
\begin{tabular}{lccccc}
\toprule
\multirow{2}{*}{\textbf{Method}} & \multicolumn{2}{c}{\textbf{FashionIQ}} & \multicolumn{3}{c}{\textbf{CIRR}}  \\
\cmidrule[0.5pt](lr){2-3} \cmidrule[0.5pt](lr){4-6} & R@10 & R@50& R@1& R@5 & R$_{\rm subset}$@1 \\
\midrule
Real  & 54.85 & 75.43 & 53.62 & 83.62 & 79.05     \\
\textbf{Pseudo} & \underline{56.57} & \underline{76.55}  & \underline{54.12} & \underline{84.27} & \underline{82.64}    \\
\bottomrule
\end{tabular}
 \label{tab:per}
\end{table}

\begin{table}[h]
\centering
\caption{Ablation studies with regard to the performance differences between pre-trained and fine-tuned models.}
\begin{tabular}{lccccc}
\toprule
\multirow{2}{*}{\textbf{Method}} & \multicolumn{2}{c}{\textbf{FashionIQ}} & \multicolumn{3}{c}{\textbf{CIRR}}  \\
\cmidrule[0.5pt](lr){2-3} \cmidrule[0.5pt](lr){4-6} & R@10 & R@50& R@1& R@5 & R$_{\rm subset}$@1 \\
\midrule
Pre-trained  & 54.42 & 75.67 & 53.22 &  83.47 & 79.16     \\
\textbf{fine-tuning} & \underline{56.57} & \underline{76.55}  & \underline{54.12} &  \underline{84.27} & \underline{82.64}    \\
\bottomrule
\end{tabular}
 \label{tab:pre}
\end{table}

\begin{figure}[t]
\centering
\includegraphics[width=\linewidth]{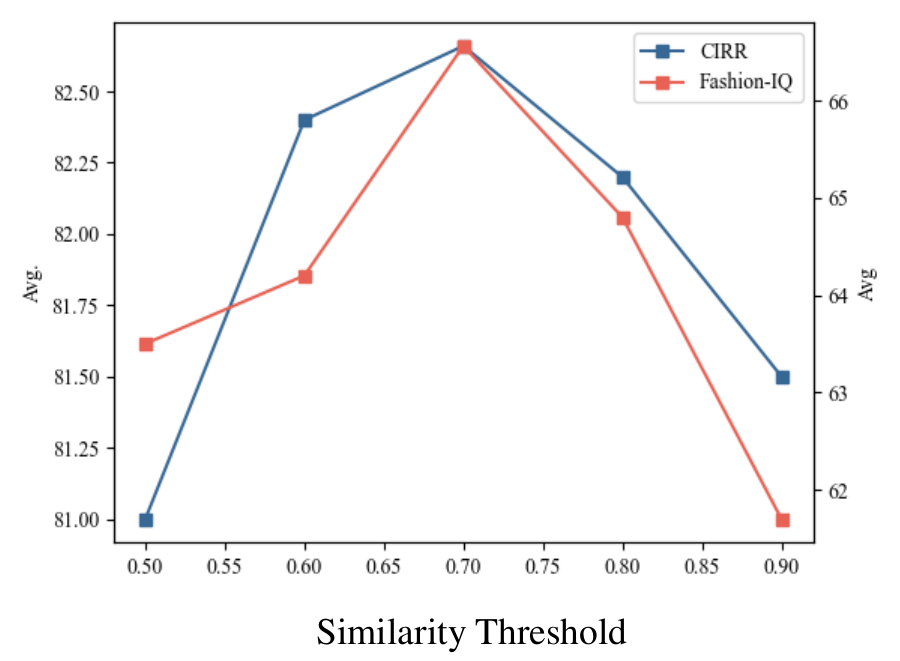}
\caption{Ablation studies in terms of average recalls with regards to different values of Similarity Threshold}
\label{fig:yuzhi}
\end{figure}

\begin{table}[h]
\centering
\caption{Ablation studies with regard to the contribution of the token merging module to retrieval performance.}
\begin{tabular}{lccccc}
\toprule
\multirow{2}{*}{\textbf{Method}} & \multicolumn{2}{c}{\textbf{FashionIQ}} & \multicolumn{3}{c}{\textbf{CIRR}}  \\
\cmidrule[0.5pt](lr){2-3} \cmidrule[0.5pt](lr){4-6} & R@10 & R@50& R@1& R@5 & R$_{\rm subset}$@1 \\
\midrule
w/o token merging    & 29.68 & 54.85 & 20.88 & 48.24  & 50.33     \\
\textbf{token merging} & \underline{56.57} & \underline{76.55}  & \underline{54.12} &  \underline{84.27} & \underline{82.64}    \\
\bottomrule
\end{tabular}
 \label{tab:tome}
\end{table}

\begin{table*}[t]
\centering
\caption {Quantitative comparison across competing methods on the Fashion-IQ validation set, where Average
indicates the average results across all the metrics in the three different classes. The best results
are marked in bold}
\begin{tabular}{lcccccccccc}
\toprule
\multirow{2}{*}{\textbf{Methods}} & \multicolumn{2}{c}{\textbf{Dress}} & \multicolumn{2}{c}{\textbf{Shirt}} & \multicolumn{2}{c}{\textbf{Top\&Tee}} &  \multicolumn{3}{c}{\textbf{Average}} \\
\cmidrule[0.5pt](lr){2-3} \cmidrule[0.5pt](lr){4-5} \cmidrule[0.5pt](lr){6-7} \cmidrule[0.5pt](lr){8-10} 
& R@10 & R@50 & R@10 & R@50 & R@10 & R@50 & R@10 & R@50 & Rmean \\
\midrule
JVSM \cite{chen2020learning} & 10.70 & 25.90 & 12.00 & 27.10 & 13.00 & 26.90 & 11.90 & 26.60 & 19.26 \\
CIRPLANT \cite{liu2021image} & 17.45 & 40.41 & 17.53 & 38.81 & 61.64 & 45.38 & 18.87 & 41.53 & 30.20 \\
TRACE w/BER \cite{jandial2022sac} & 22.70 & 44.91 & 20.80 & 40.80 & 24.22 & 49.80 & 22.57 & 46.19 & 34.00 \\
VAL \cite{chen2020image} w/GloVe & 22.53 & 44.00 & 22.38 & 44.15 & 27.53 & 51.68 & 24.15 & 46.61 & 35.38 \\
MAAF \cite{dodds2020modality} & 23.80 & 48.60 & 21.30 & 44.20 & 27.90 & 53.60 & 24.30 & 48.80 & 36.60\\
CurlingNet \cite{yu2020curlingnet} & 26.15 & 53.24 & 21.45 & 44.56 & 30.12 & 55.23 & 25.90 & 51.01 & 34.36\\
RTIC-GCN \cite{shin2021rtic} & 29.15 & 54.04 & 23.79 & 47.25 & 31.61 & 57.98 & 28.18 & 53.09 & 40.64\\
CoSMo \cite{lee2021cosmo} &  25.64 & 50.30 & 24.90 & 49.18 & 29.21 & 57.46 & 26.58 & 52.31 & 39.45\\
ARTEMIS \cite{delmas2022artemis} & 27.16 & 52.40 & 21.78 & 43.64 & 29.20 & 53.83 & 26.05 & 50.29 & 38.04\\ 
DCNet \cite{kim2021dual} & 28.95 & 56.07 & 23.95 & 47.30 &30.44 & 58.29 & 27.78 & 53.89 & 40.84\\
SAC w/BERT \cite{jandial2022sac} & 26.52 & 51.01 & 28.02 & 51.86 & 32.70 & 61.23 & 29.08 & 54.70 & 41.89\\
FashionVLP \cite{goenka2022fashionvlp} & 32.42 & 60.29 & 31.89 & 58.44 & 38.51 & 68.79 & 34.27 & 62.51 & 48.39\\
LF-CLIP(Combiner) \cite{baldrati2022effective} & 31.63 & 56.67 & 36.36 & 58.00 & 38.19 & 62.42 & 35.39& 59.03& 47.21\\
LF-BLIP \cite{levy2024data} & 25.31 & 44.05 & 25.39 & 43.57 & 26.54 & 44.48 & 25.75 & 43.98 & 34.88\\
CASE \cite{zhu2023amc} & 47.44 & 69.36 & 48.48 & 70.23 & 50.18 &  72.24 & 48.79 & 70.68 & 59.74\\
AMC \cite{ventura2024covr} & 31.73 & 59.25 & 30.67 & 59.08 & 36.21 & 66.06 & 32.87 & 61.64 & 47.25\\
CoVR-BLIP \cite{baldrati2022conditioned} & 44.55 & 69.03 & 48.43 & 67.42 & 52.60 & 74.31 & 48.53 & 70.25 & 59.39\\
CLIP4CIR \cite{liu2024bi} & 33.81 & 59.40 & 39.99 & 60.45 & 41.41 & 65.37 & 38.32 & 61.74 & 50.03\\
BLIP4CIR+Bi \cite{han2023fame} & 42 09 &  67.33 & 41.76 & 64.28 & 46.61 & 70.32 & 43.49 & 67.31 & 55.04\\
FAME-ViL \cite{wen2023target}& 42.19 & 67.38 & 47.64 & 68.79 & 50.69 & 73.07 & 46.84 & 69.75 & 58.29\\
TG-CIR \cite{jiang2023dual} & 45.22 & 69.66 & 52.60 & 72.52 & 56.14 & 77.10 & 51.32 & 73.09 & 58.05\\
Re-ranking \cite{liu2023candidate} & 48.14 & 71.43 & 50.15 & 71.25 & 55.23 & 76.80 & 51.17 & 73.13 & 62.15\\
CompoDiff \cite{gu2023compodiff} & 40.65 & 57.14 & 36.87 & 57.39 & 43.93 & 61.17 & 40.48 & 58.57 & 49.53\\
SPRC \cite{bai2023sentence} & 49.18 & 72.43 & 55.64 & 73.89 & 59.35 & 78.58 & 54.92 & 74.97 & 64.85\\
\midrule
TMCIR (Ours) & \textbf{50.67} & \textbf{73.86} & \textbf{59.12} & \textbf{76.34} & \textbf{59.93} & \textbf{79.46} & \textbf{56.57} & \textbf{76.55} & \textbf{66.56}\\
\bottomrule
\end{tabular}
\label{tab:fiq}
\end{table*}

\begin{table*}[ht]
\centering
\caption{Quantitative comparison across competing methods on the CIRR test set, where Avg. indicates the
average results across all the metrics in the three different settings.
The best results are marked in bold}
\begin{tabular}{lccccccccc}
\toprule
\multirow{2}{*}{\textbf{Methods}} & \multicolumn{4}{c}{\textbf{Recall@K}} & \multicolumn{3}{c}{\textbf{R$_{\rm subset}$@K}} & \multirow{2}{*}{\textbf{Avg.}} \\
\cmidrule[0.5pt](lr){2-5} \cmidrule[0.5pt](lr){6-8} 
& K=1 & K=5 & K=10 & K=50 & K=1 & K=2 & K=3 \\
\midrule
TIRG \cite{vo2019composing} & 14.61 & 48.37 & 64.08 & 90.03 & 22.67 & 44.97 & 65.14 & 35.52 \\
TIRG+LastConv \cite{vo2019composing} & 11.04 & 35.68 & 51.27 & 83.29 & 23.82 & 45.65 & 64.55 & 29.75 \\
MAAF \cite{dodds2020modality} & 10.31 & 33.03 & 48.30 & 80.06 & 21.05 & 42.24 & 61.60 & 27.08 \\
MAAF-BERT \cite{dodds2020modality} & 10.12 & 33.10 & 48.01 & 80.57 & 22.08 & 42.41 & 62.14 & 27.57 \\
MAAF-IT \cite{dodds2020modality} & 9.90 & 32.86 & 48.83 & 80.27 & 21.17 & 40.91 & 60.91 & 27.02 \\
MAAF-RP \cite{dodds2020modality} & 10.22 & 33.32 & 48.68 & 81.84 & 21.41 & 42.04 & 61.60 & 27.37 \\
CIRPLANT \cite{liu2021image} & 19.55 & 52.55 & 68.39 & 92.38 & 39.20 & 63.03 & 79.49 & 45.88 \\
ARTEMIS \cite{delmas2022artemis} & 16.96 & 46.10 & 61.31 & 87.73 & 39.99 & 62.20 & 75.67 & 43.05 \\
LF-BLIP \cite{baldrati2022effective} & 20.89 & 48.07 & 61.16 & 83.71 & 50.22 & 69.39 & 86.82 & 60.58 \\
LF-CLIP \cite{baldrati2022effective} & 33.59 & 65.35 & 77.35 & 95.21 & 62.39 & 79.74 & 92.02 & 72.53 \\
CLIP4CIR \cite{baldrati2022conditioned} & 38.53 & 69.98 & 81.86 & 95.93 & 68.19 & 86.31 & 94.17 & 69.09 \\
BLIP4CIR+Bi \cite{liu2024bi} & 40.15 & 73.08 & 83.88 & 96.27 & 72.10 & 90.16 & 95.93 & 72.59 \\
CompoDiff \cite{gu2023compodiff} & 22.35 & 54.36 & 73.41 & 91.77 & 35.84 & 58.21 & 76.60 & 29.10 \\
CASE \cite{levy2024data} & 48.00 & 79.11 & 87.25 & 97.57 & 75.88 & 94.67 & 96.00 & 77.50 \\
CASE Pre-LaSCo.CaT \cite{levy2024data} & 49.35 & 80.02 & 88.75 & 97.47 & 76.48 & 95.03 & 95.71 & 78.25 \\
TG-CIR \cite{wen2023target} & 45.25 & 78.29 & 87.16 & 97.30 & 72.84 & 89.25 & 95.13 & 75.57 \\
DRA \cite{jiang2023dual} & 39.93 & 72.07 & 83.83 & 96.43 & 71.04 & 91.43 & 94.72 & 71.55 \\
CoVR-BLIP \cite{ventura2024covr} & 49.69 & 78.60 & 86.77 & 94.31 & 75.01 & 91.07 & 93.16 & 80.81 \\
Re-ranking \cite{liu2023candidate} & 50.55 & 81.75 & 89.78 & 97.18 & 80.04 & 94.29 & 96.80 & 80.90 \\
SPRC \cite{bai2023sentence} & 51.96 & 82.12 & 89.74 & 97.69 & 80.65 & 92.31 & 96.60 & 81.39 \\
CIR-LVLM \cite{sun2024leveraging} & 53.64 & 83.76 & 90.60 & 97.93 & 79.12 & 92.33 & 96.67 & 81.44 \\
\midrule
TMCIR (Ours) & \textbf{54.12} & \textbf{84.27} & \textbf{91.06} & \textbf{98.43} & \textbf{82.64} & \textbf{92.45} & \textbf{96.77} & \textbf{82.66} \\
\bottomrule
\end{tabular}
\label{tab:cirr}
\end{table*}

\subsection{Comparative study}

\textbf{Results on Fashion-IQ}  
Table \ref{tab:fiq} summarizes the evaluation of various competing methods on the \textbf{Fashion-IQ} dataset. As observed from Table \ref{tab:fiq}, our proposed method, TMCIR, achieves the highest recall rates across all eight evaluation metrics for the three clothing categories. Specifically, compared to the second-best method SPRC, TMCIR improves the R@10 metric from 54.92 to 56.57. More importantly, in the case of the text-to-image retrieval paradigm embodied by SPRC, our method demonstrates a significant enhancement on the "Shirt" category, with R@10 increasing from 55.64 to 59.12 and R@50 improving from 73.89 to 76.34. This improvement is primarily attributable to the fact that SPRC abstracts and converts a portion of the reference image information into textual prompts, and relies solely on a text-based retrieval model for matching. Such an indirect utilization of visual information is more prone to losing certain fine-grained visual details of the original reference image when compared to methods that directly fuse visual features. In contrast, our TMCIR method adaptively merges visual tokens and textual tokens via the token merging module, thereby producing a richer and more fine-grained multimodal fused representation. This enables the capture and exploitation of essential visual details in the reference image (\eg, specific patterns, textures, or styles within the "Shirt" category), resulting in improved retrieval performance.

\textbf{Results on CIRR}  
We further evaluated the effectiveness of our method on the more general CIRR dataset, with the results presented in Table \ref{tab:cirr}. Compared to the existing text prompt-based method SPRC, our approach consistently demonstrates improvements on CIRR. For instance, Recall@1 increases from 51.96 to 54.12, Recall@5 from 82.12 to 84.27, Recall@10 from 89.74 to 91.06, and Recall@50 from 97.69 to 98.43. Clearly, although the sentence-level prompts learned in SPRC enhance the relative textual information, they are unable to fully encode the subtle visual features present in the reference image. This shortcoming leads to suboptimal performance in categories requiring precise visual matching (\eg, "Shirt"), thereby resulting in a marked performance gap when compared to our method.

In terms of the Recallsubset@K metrics, our TMCIR also outperforms SPRC, for example: Recall@1 improves from 80.65 to 82.64, Recall@2 from 92.31 to 92.45, and Recall@3 from 96.60 to 96.77.

Moreover, when compared to the strongest CIR-LVLM baseline, our TMCIR still attains the highest recall rates. For example, on the three Recallsubset@K metrics, SPRC yields increases of 3.52, 0.12 and 0.10, respectively, whereas our TMCIR, by introducing the similarity-based token fusion module and task-specific encoder fine-tuning, achieves significantly more effective multimodal information fusion. This enables a better understanding of complex composed image queries and, as a result, substantially enhances the accuracy of composed image retrieval.

\subsection{Ablation Study}

\textbf{Pseudo-target Images vs. Real Target Images}
We investigate the impact of utilizing pseudo-target images generated by a diffusion model versus real target images from the Composed Image Retrieval (CIR) dataset during the encoder fine-tuning phase. Specifically, two training configurations are designed: one employs real target images as supervision signals, while the other uses pseudo-target images conditioned on the reference image and relative description. Experimental results on the CIRR dataset, as shown in Table ~\ref{tab:per}, indicate that pseudo-target images more accurately capture the intended modifications, providing cleaner and more directive supervision. This enhances the alignment and robustness in subsequent token fusion stages. In contrast, real target images may introduce background noise and other distractions, hindering the model's ability to capture fine-grained semantic differences, thereby affecting overall retrieval performance.

\textbf{Impact of Similarity Threshold in Token Merging}
To assess the effect of the similarity threshold $\tau$ in the token merging module on cross-modal fusion and overall CIR performance, we experiment with various threshold settings (e.g., $\tau = 0.5$, $0.7$, $0.9$). As illustrated in Table ~\ref{fig:yuzhi}, an optimal threshold (e.g., $\tau = 0.7$) effectively filters out noisy tokens while retaining sufficient informative ones, leading to higher semantic consistency and discriminative power in the fused tokens. A lower threshold (e.g., $\tau = 0.5$) admits excessive noisy tokens, degrading retrieval accuracy, whereas a higher threshold (e.g., $\tau = 0.9$) results in the loss of valuable information. Eliminating the matching strategy altogether fails to leverage similarity information, causing significant performance drops due to ineffective cross-modal alignment.

\textbf{Pre-trained Models vs. Task-specific Fine-tuning}
To validate the efficacy of task-specific fine-tuning in CIR, we compare the performance of the pre-trained CLIP encoder against its fine-tuned counterpart. In the "no fine-tuning" setup, the CLIP model directly extracts representations for the reference image and relative description, followed by token fusion and similarity computation. In the "fine-tuning" setup, we construct image-text pairs using pseudo-target images and apply contrastive loss to fine-tune both the visual and textual encoders of CLIP, achieving precise cross-modal token alignment in a shared embedding space. The experimental results are shown in Table ~\ref{tab:pre}.Experimental results demonstrate that the fine-tuned model exhibits significant improvements across all metrics. For instance, Recall@1 increases by approximately $0.89$\%, indicating that task-specific fine-tuning effectively mitigates cross-modal alignment issues inherent in pre-trained models under specific data distributions. Moreover, the fine-tuned model shows enhanced consistency and robustness in token representations, better capturing the semantic modifications between the reference image and relative description. This ablation study underscores the importance of employing contrastive learning strategies to fine-tune pre-trained models for improved CIR performance.

\textbf{Contribution of the Token Merging Module}
To evaluate the effectiveness of our proposed token merging module, we design two model variants: one incorporates the token merging strategy, which adaptively integrates cross-modal features by computing token similarities and performing weighted fusion; the other bypasses token matching and fusion, directly utilizing the original token sequences from the visual and textual encoders for subsequent average pooling and fully connected mapping.The experimental results are shown in Table ~\ref{tab:tome}. Experimental results reveal that the model without the token merging strategy exhibits noticeable declines in cross-modal image retrieval tasks, particularly in fine-grained semantic alignment and overall retrieval accuracy (e.g., Recall@K metrics). This validates the critical role of the token merging module in suppressing noise, enhancing inter-token mutual information, and improving the consistency of final cross-modal feature representations. Further analysis of feature distributions before and after token fusion indicates that the token merging module significantly reduces distributional discrepancies between visual and textual tokens, thereby better supporting subsequent image retrieval tasks.
\section{Conclusion}
Addressing the challenge of biased feature fusion in Composed Image Retrieval (CIR), we introduced TMCIR. Our framework leverages \textbf{Intent-Aware Cross-Modal Alignment (IACMA)}, using diffusion-generated pseudo-target images for cleaner encoder fine-tuning, and \textbf{Adaptive Token Fusion (ATF)}, which merges tokens based on similarity and position to balance modalities. Extensive experiments demonstrate that TMCIR significantly outperforms state-of-the-art methods on the Fashion-IQ and CIRR benchmarks. By effectively preserving visual details while accurately capturing textual modification intent, TMCIR offers a more robust and precise solution for CIR. 











\end{document}